# LoST? Appearance-Invariant Place Recognition for Opposite Viewpoints using Visual Semantics

Sourav Garg[†], Niko Suenderhauf and Michael Milford
Australian Centre for Robotic Vision, Robotics and Autonomous Systems,
School of Electrical Engineering and Computer Science, Queensland University of Technology, Brisbane
[†]Email: `sourav.garg@hdr.qut.edu.au`

*Abstract*—Human visual scene understanding is so remarkable that we are able to recognize a revisited place when entering it from the opposite direction it was first visited, even in the presence of extreme variations in appearance. This capability is especially apparent during driving: a human driver can recognize where they are when travelling in the reverse direction along a route for the first time, without having to turn back and look. The difficulty of this problem exceeds any addressed in past appearance- and viewpoint-invariant visual place recognition (VPR) research, in part because large parts of the scene are not commonly observable from opposite directions. Consequently, as shown in this paper, the precision-recall performance of current state-of-the-art viewpoint- and appearance-invariant VPR techniques is orders of magnitude below what would be usable in a closed-loop system. Current engineered solutions predominantly rely on panoramic camera or LIDAR sensing setups; an eminently suitable engineering solution but one that is clearly very different to how humans navigate, which also has implications for how naturally humans could interact and communicate with the navigation system. In this paper we develop a suite of novel semantic- and appearance-based techniques to enable for the first time high performance place recognition in this challenging scenario. We first propose a novel Local Semantic Tensor (LoST) descriptor of images using the convolutional feature maps from a state-of-the-art dense semantic segmentation network. Then, to verify the spatial semantic arrangement of the top matching candidates, we develop a novel approach for mining semantically-salient keypoint correspondences. On publicly available benchmark datasets that involve both 180 degree viewpoint change and extreme appearance change, we show how meaningful recall at 100% precision can be achieved using our proposed system where existing systems often fail to ever reach 100% precision. We also present analysis delving into the performance differences between a current and the proposed system, and characterize unique properties of the opposite direction localization problem including the metric matching offset.

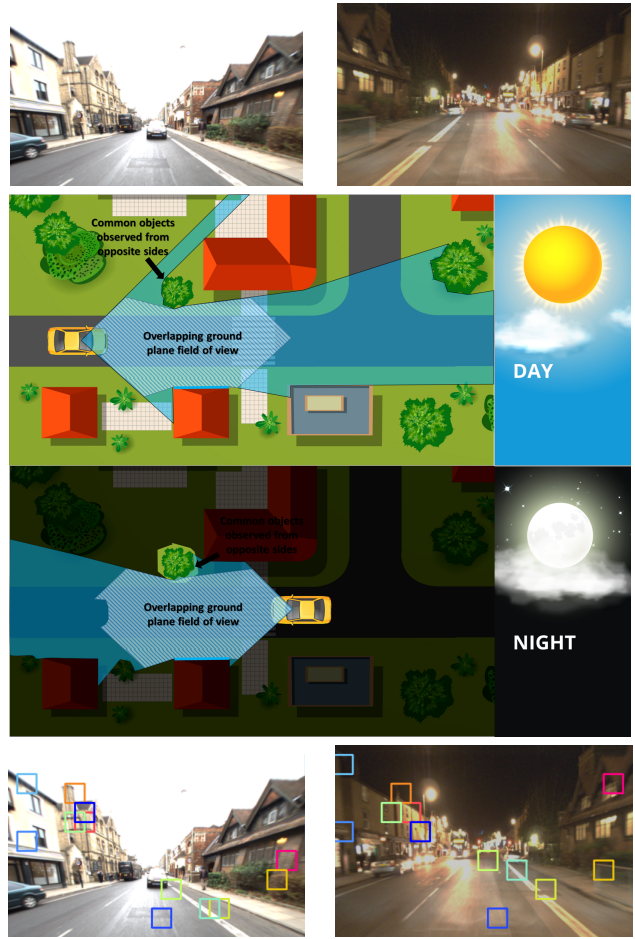

Fig. 1. Recognizing a place that is revisited from the opposite direction, especially when environmental conditions are not the same, is a challenging problem. The top rows illustrate the amount of visual overlap available for matching images in such scenarios. The bottom row shows semantic *keypoint* correspondences (of the same color) using the proposed LoST-X framework (The query image is flipped for spatial layout verification).

## I. INTRODUCTION

Semantic scene understanding, though theorized decades ago [4, 32], has only recently become a success with the advent of deep learning methods [20]. The use of visual semantics not only enables meaningful interaction with practical applications, but also aids in solving more complex problems that require human-like interpretation of the environment. One such problem pertains to visual place recognition under viewpoint variations as extreme as front- and rear-view image matching. This problem, encountered regularly by humans when driving, is particularly challenging because only a subset of the scene is commonly observable from both directions, and becomes even more challenging when the appearance of the environment varies due to day-night cycles, weather or seasonal conditions.

Current state-of-the-art visual place recognition methods employ deep-learnt techniques [30, 1] to primarily solve challenges related to variations in appearance; the extent of viewpoint-invariance is often limited to variations in lateral displacement, orientation, and scale relative to the reference

6-DoF camera pose in the real world. The multi-view matching problem is generally approached with sensors-based engineering solutions using panoramic cameras [2] and LIDARs [41]. Dealing with 180 degree viewpoint change using only a limited field-of-view camera under extreme appearance variation is an unsolved problem. We show here that current state-of-the-art systems, even with multi-frame filtering, achieve performance levels far below the usable values for a practical application. To address this challenge, we propose a place recognition system that utilizes semantic scene information to deal with the viewpoint and appearance variations, and demonstrate its ability to achieve meaningful recall at 100% precision. Our approach is also motivated by the potential for a semantically-enabled system to be more useful for human communication and interaction with robots and vehicles.

In particular, we make the following core contributions:

- A novel visual place recognition pipeline combining semantic- and appearance-based global descriptors with spatially-consistent local keypoint correspondences.
- A novel image descriptor, referred to as LoST, generated from the semantic labels and convolutional feature maps of the input image.
- A novel method to mine and filter *semantically-salient* keypoint correspondences from the matching pair of images.
- The first demonstration of a single camera, limited field of view system that is able to achieve meaningful recall at 100% precision despite the combination of both opposing viewpoints and large appearance change.

The paper proceeds as follows: Section II reviews the literature for related work and provides a background for the problem; Section III describes the proposed pipeline; Section IV describes the experimental setup; Section V shows the results with comparisons to state-of-the-art methods and Section VI provides conclusions from the findings of our study and areas of future work. The source code for this paper is available online[1].

## II. LITERATURE REVIEW

In the last decade, visual place recognition research has progressed through Bag of Visual Words (BoVW) based methods like FAB-MAP [7] and global image descriptor based sequence searching [26, 29] to deep-learnt robust place representations [1, 5]. The deep-learnt networks have been used in several ways for visual place recognition, ranging from ConvNet Landmarks [36], deconvolution network [27], direct off-the-shelf convolutional feature descriptions [35], deep feature pooling methods based on sum [3], cross-convolution [22], VLAD [42] to end-to-end training using NetVLAD [1].

The deep-learnt representations of places often use a global image descriptor for robust matching but do not retain the spatial layout of the images which is crucial to deal with *perceptual aliasing*. Towards this end, the authors in [5] mine distinctive landmarks from within the images but require a visual vocabulary to encode landmark regions. [38] uses region-wise approximate integral max-pooling to retrieve best matches, however object localization requires a correspondence search procedure. [39] finds dense visual correspondences using a spatial feature pyramid and [19] learns local self-similarity (LSS) for the same purpose, but both focus primarily on objects and not places. [9] uses spatial softmax to learn high-activation spatial features for visuomotor learning. The authors in [33, 14] train a CNN for determining correspondences between images by geometric matching. The authors in [15] use a Hypercolumns-based pixel-wise representation for fine-grained localization of keypoints and object parts. However, most of these methods do not leverage the *semantic correspondence* information pre-coded in the convolutional neural networks.

The rapid advancement in the field of dense pixel-wise semantic segmentation of images [23, 21] using deep convolutional neural networks has paved the way for leveraging semantic scene information in other domains of computer vision research, like fine-grain object segmentation [15], SLAM [8], mapping [37] etc. Similarly, it has a key role to play in visual place recognition and localization, especially for matching images with viewpoint-variation as extreme as front-view vs rear-view.

The use of semantics for visual place recognition has received limited attention. The authors in [30] use semantic masking of appearance-invariant classes like buildings and develop an aggregated convolutional feature descriptor, but this requires environment-sepcific training. [11] uses semantic place categorization to improve place recognition for seamlessly navigating through diverse environment types, but only uses environment-related visual semantics. The authors in [17] use geometric-pairs of semantic landmarks suited for roadways to deal with environmental variations. While most of these methods use semantic information only for dealing with appearance variations, authors in [13] proposed a multi-view place localization system, termed as X-view, that deals with extreme viewpoint variations, however, the method relies solely on the semantic labels without leveraging the appearance cues.

The research presented in this paper is motivated by the belief that achieving human-like place recognition across opposing viewpoints and extreme appearance change requires a deeper semantic representation of the environment. In this work we set out to achieve this by introducing visual semantics into the place recognition pipeline for semantically describing places both at image- and pixel-level, as described in the following section.

## III. PROPOSED APPROACH

The proposed method builds upon the traditional hierarchical procedure [7, 28] of finding top candidates with a global database-wide search and then verifying the spatial layout of the candidate pairs to select the best match. Additionally, we introduce the visual semantic information at both stages; both in generating semantically-aggregated descriptions of a place and mining semantically-consistent *keypoint* correspondences; and in turning a dense pixel-wise semantic segmentation network into a visual place recognition pipeline.

---
[1]https://github.com/oravus/lostX

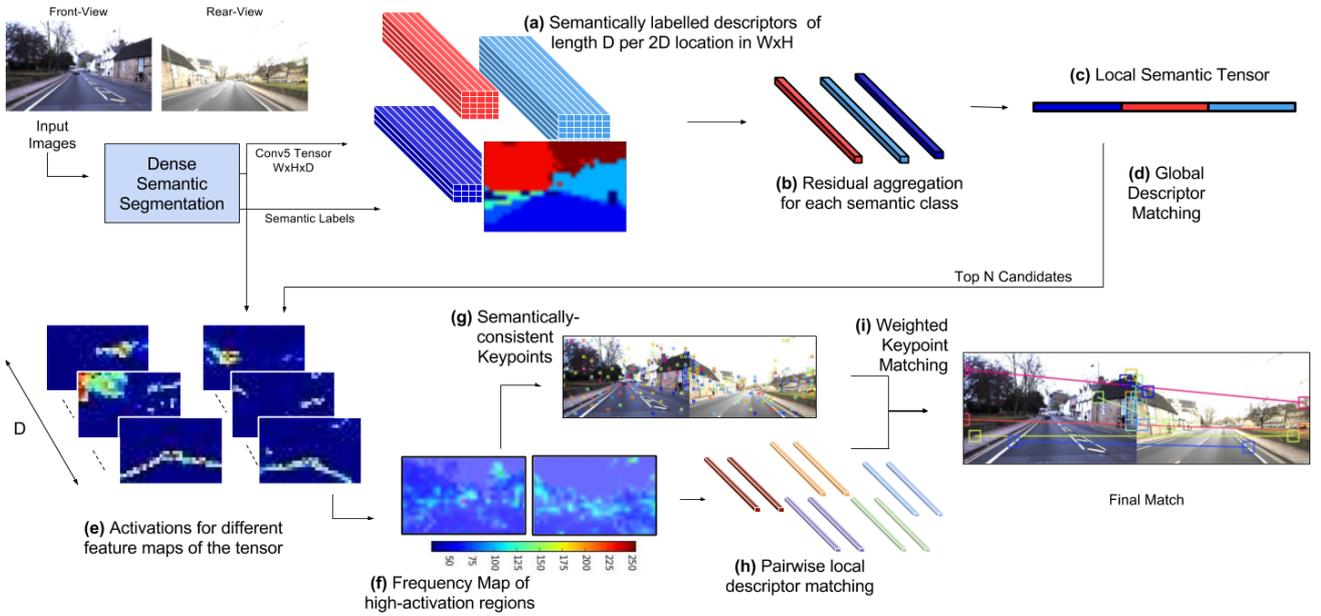

Fig. 2. Proposed Approach

*A. Local Semantic Tensor*

The representation of images using different methods of convolutional feature pooling has been widely addressed in literature [22]. In most of these scenarios, the semantic regions within the images are not explicitly targeted, rather different regions are extracted from higher-order layers of CNN that tend to capture visual semantics implicitly. During the matching stage, it adds the overhead of environment-specific training [30], cross-matching [38] or vocabulary learning [5]. Here, we develop a method to semantically pool features using the semantic label scores and the convolutional feature maps of a dense semantic segmentation network.

For this task, we use a state-of-the-art semantic segmentation network RefineNet [21] trained on the Cityscapes Dataset [6]. The convolutional feature maps are extracted from the conv5 layer of the network. These feature maps form a tensor of size $W \times H \times D$, where $W$, $H$ and $D$ are the width, height and depth of the feature maps. In this case, the W and H are $1/32^{th}$ of the input image, $D$ is 2048. Similarly, the semantic label scores so obtained have W and H as $1/4^{th}$ of the input image [2] and $D$ is 20 which corresponds to the number of semantic categories in the Cityscapes Dataset.

Keeping in mind the noise associated with the dense semantic labels and the success of image description methods like VLAD (Vector of Locally Aggregated Descriptors) [18], we define a semantic descriptor $L$ for an image, referred to as Local Semantic Tensor (LoST), using the convolutional feature maps and semantic label probability as shown in Figure 2 (a)-(c)

$$L_s = \sum_i^N m_{is} |\mu_s - x_i| \quad (1)$$

$$\mu_s = \frac{\sum_i^N \{x_i \mid l_i = s\}}{\sum_i^N \{i \mid l_i = s\}} \quad (2)$$

$$l_i = \arg\max_s m_{is} \quad (3)$$

where $l_i$ is the semantic label of a $D$-dimensional descriptor $x_i$ at location $i$ within the feature map as shown in Figure 2 (a). $\mu_s$ is the mean descriptor with respect to the semantic category $s$. While the mean is computed using the most likely labels for the pixels, the image descriptor $L_s$ is computed using the probability $m_{is}$ of a pixel location $i$ to belong to a semantic class $s$. The $m_{is}$ is computed by L1-normalization of the label scores tensor across the dimension $D$. Each semantic descriptor $L_s$ is essentially an aggregation of the residual description from that particular semantic class $s$ and the noisy contributions from the remaining classes, weighted by their semantic label probability scores. The final image descriptor $L$ is a concatenation of L2-normalized $L_s$ using only three semantic classes, that is, road, building and vegetation. Each image descriptor for both reference and query database undergoes normalization as described here [12] before descriptor matching. For the reference database, as available beforehand, we modify the Equation 2 to replace the mean computed from a single image by the mean computed over a sliding window of 15 frames centered at the given reference image.

[2] The semantic score tensor is resized to conv5's resolution when used in conjunction with its feature maps.

*Candidate Search:* A query image descriptor is matched with all the images in the database using the cosine distance. The top 10 candidates with the lowest distance to the query are passed to the next step as described below for determining the final match.

### B. Correspondence Extraction

The traditional image matching methods based on local point features [7] first extract salient keypoints and then use a robust descriptor [24] to match images using RANSAC [10]. On the other hand, the deep-learnt image representations used for place recognition are generally global image descriptors [1] that have implicit information about the local visual landmarks within the image. However, this local saliency information can be extracted from the network in different ways by leveraging the high activation regions within the convolutional feature maps. These regions are often used for constructing a more robust representation [5, 38].

We propose another perspective to these *maximally-activated* locations, referred to as *keypoints* in our case, extracted from each of the feature maps of higher-order convolutional layers of the network. It is well-understood now that higher-order layers of the convolutional neural networks capture visual semantics, and that different convolutional filters for one such layer (say conv5) will repeatedly detect the same semantic concepts from perceptually similar images [44, 43]. Therefore, it can be safely assumed that there exists a correspondence between the convolutional feature maps of two such images and also that these correspondences can be mapped to high-activation *keypoint* locations in the feature maps. This is shown in Figure 2 (e)-(f). For a given matching pair of images, we can extract a total of $D$ (depth/number of feature maps) *keypoint* correspondences from a higher-order layer of the network.

For an identical or closely matching pair of images, these correspondences would probably be near to perfect, but for a practical scenario there would be many false correspondences due to the following reasons: 1) the convolutional filters inherently trained to detect a particular concept/feature which is not present in the given image may fire at random locations, for example, a traffic-light sensitive filter applied to an indoor image, 2) multiple instances of a concept within a single image, for example, windows or trees, may cross-correspond, 3) the dynamic objects like cars, pedestrians etc. appear at random locations in the image and can cause false correspondence.

It is worth noting that the actual number of correspondences $D$ can be significantly larger than the resolution ($W \times H$) of the feature maps of that layer. This is true in general for many network architectures as well as the one considered in our work. For example, conv5 of AlexNet [20] has 384 feature maps as compared to its resolution of $7 \times 7$ and ResNet - 101 [16] based RefineNet in our case has these values 2048 and $31 \times 20$ respectively. Figure 2 (f) shows a frequency map of high activation regions within the image as counted across all the feature maps. It can be observed that a relatively small number of *keypoint* locations tend to trigger activations in a large number of feature maps.

### C. Spatial Layout Verification

The global spatial layout verification of the candidate pairs of images is achieved through weighted keypoint matching. This necessitates limiting the total number of correspondences as well as avoiding false correspondences, which is achieved using a prior filtering based on semantic label consistency as described below:

*1) Semantic Label Consistency:* The correspondence formulation as described in the previous section is applicable to any pre-trained convolutional neural network architecture. Additionally, as in our case, the semantic segmentation network provides pixel-wise semantic labels. We use these labels (after resizing them to the resolution of conv5 layer $31 \times 20$) to obtain a semantic label for each of the *keypoint* locations within the feature maps. Finally, for each corresponding pair of *keypoints* the semantic labels are verified and those with similar labels are retained. This step filters out more than half of the correspondences, especially those related to random activations caused by convolutional filters that detect concepts/features not available in a particular scene.

The semantically-consistent correspondences are further filtered by a neighborhood density test such that within a $3 \times 3$ neighborhood of a *keypoint* only one correspondence is retained. This filtering helps in removing redundant correspondences that belong to the same semantic class within a very small region, for example, multiple *keypoints* on a glass pane of an office building. The cross-correspondence due to multiple instances of a semantic class can not be filtered by this method and a worst case scenario may occur when cross-correspondences survive, filtering out the true correspondences. Therefore, pairwise descriptor matching of such correspondences is required to differentiate between them on the basis of their appearance, as described in the next section.

*2) Weighted Keypoint Matching:* The remaining false correspondences are dealt with by using a pairwise descriptor matching of each correspondence that enables differentiation between the fine-grained appearance of the *keypoints*. Therefore, for the remaining corresponding *keypoints*, say $M$, $D$-dimensional descriptors $x_k$ and $x_{k'}$ at the given corresponding *keypoint* locations $k$ and $k'$ are extracted from the conv5 layer of the network as shown in Figure 2 (g)-(h). Further, the x-coordinate locations $p_k$ and $p_{k'}$ (flipped from left to right for rear-view images) associated with each of the corresponding *keypoints* are obtained. These *keypoint* locations are then matched by calculating the weighted Euclidean distance:

$$r_c = \sqrt{\sum_{k,k'}^{M} w_{kk'}(p_k - p_{k'})^2} \quad (4)$$

where $w_{kk'}$ is the cosine similarity between the corresponding descriptors $x_k$ and $x_{k'}$ normalized across all pairs $M$. $r_c$ is the matching score for a shortlisted candidate $c$; the candidate with the lowest score is considered the final match.

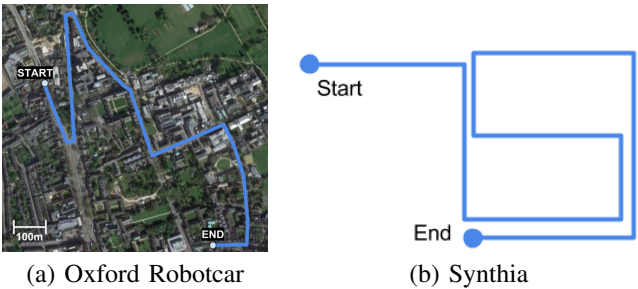

(a) Oxford Robotcar  (b) Synthia

Fig. 3. (a) Aerial View of the trajectory for the Oxford Robotcar dataset. (b) Approximate route for the Synthia dataset. Source: Google Maps

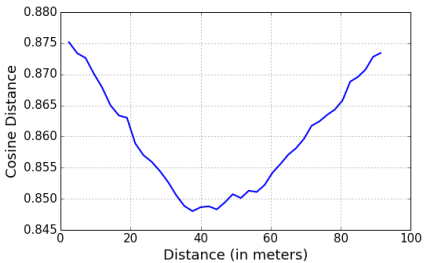

Fig. 4. Average cosine distance between a pair of front-view and rear-view images plotted against the average distance between them. This curve shows that on an average there is approximately 40 meters of *visual offset* between a matching pair of front- and rear-view images.

### D. Image Sequence Matching

In place recognition research for mobile robots or vehicles, it is a common practice to leverage the temporal sequence information [7, 29] inherent in the problem. Given the difficulty (as we show in Section V B) of the problem addressed here: a limited amount of visual overlap (that itself can have significant appearance variation) that is available for matching two images with opposite viewpoints, as shown in Figure 1, we also leverage temporal sequence information here. We use a slight variant of the publicly available OpenSeqSLAM [26] with an induced ability to accumulate sequence scores for only the top matching candidates generated using the LoST-X framework (apart from the traditional global search for descriptor-only methods). Consequently the sequence-based method is not a claimed contribution of this work but rather a standard practice enhancement, and as the results show, sequence-enhancing the current state-of-the-art systems alone is not sufficient to achieve practical recall performance at 100% precision. Other sequence-based methods [31, 29, 40] could also be used.

## IV. EXPERIMENTAL SETUP

### A. Datasets

We used two different publicly available benchmark datasets that comprise image data for both varying viewpoints (that is, front- and rear-view) as well as varying environmental conditions (that is, time of day, season etc.), as described below:

*1) Oxford Robotcar:* The Oxford Robotcar Dataset [25] comprises various traverses of Oxford city from different times of day, seasons and weather conditions. We use three different traverses from the Oxford Robotcar dataset referred to as Overcast Autumn, Night Autumn and Overcast Summer [3]. We use the rear-view imagery from the Overcast Autumn Traverse to match it with the front-view imagery from all the three traverses. We use the initial 2 km of the original traverse for all the experiments. GPS data is used to sample the images at a constant distance of approximately 2 meters that leads to around 600-900 image frames in all the traverses. Figure 3(a) shows an aerial view of the trajectory for the traverses used in our experiments.

*2) Synthia Dataset:* The Synthia Dataset [34] is a collection of various sequences of images from a virtual world with varying environmental conditions as well as environment types. We used the front- and rear-view images from the *Dawn* and *Fall* traverses of *Sequence-02* (New York-like city). Each of these traverses is approximately 1.5 km in length with 941 and 742 frames in the *Dawn* and *Fall* sequences, respectively. Figure 3(b) provides an approximate route of the traverses.

### B. Ground Truth

The calculation of ground truth for front- vs rear-view matching is non-trivial. Unlike the front-view only scenario, where every image in one traverse can be closely associated to its counterpart in the second traverse, ground truth for front-rear matching depends on the persistence of salient visual landmarks in the field of view. For example, a building on the left with a span of 20 meters will generate the best matching score when front- and rear-view images are 20 meters apart, that is, the cameras are placed at opposite ends of the building. This distance, referred to as *visual offset* from here, depends on the type of visual landmarks, their size and position as well as the camera's field-of-view.

Figure 4 shows the average cosine distance between a given pair of front- and rear-view images plotted against the average distance between them (calculated through GPS information from Oxford dataset). It can be noted that minima of this curve lies near to 40 meters which means that a localization error of 10 meters as per the traditional front-front place recognition is theoretically equivalent to 50 meters of localization error for front-rear matching. However, the 40 meters value is possibly inflated because the curve in Figure 4 was generated by assuming that the matching algorithm (LoST-X) perfectly captures the variations in image matching as the distance from the true match varies and secondly that for each image there exists a correct match (which is not true when vehicle takes a turn). In practice, we found 30 meters to be the average *visual offset* as observed visually and confirmed through GPS information for a handful of image pairs.

[3]Originally 2014-12-09-13-21-02, 2014-12-10-18-10-50, 2015-05-19-14-06-38 respectively in [25]

## C. Evaluation Criteria

We primarily used Precision-Recall curves for performance evaluation. The matching score values (Equation 4) were directly used as a threshold to generate P-R curves. We found that in order to achieve a practically usable system, the localization radius required to classify a match as a true positive was 80 meters, as calculated from the GPS location of the given pair of images. However, as described in the previous section, there exists a *visual offset* of 30-40 meters between the ground truth match and the actual distance between the matching pair of images (Figure 4). This implies that the 80 meters of localization radius is equivalent to 40-50 meters of accuracy if it were to be the traditional front-view only place recognition. The sequence length used was 100 meters for all the experiments.

For our later analysis of *single-image* matching, we use a stricter value of 40 meters which is equivalent to 0-10 meters of front-view only scenario. Furthermore, the max-F1 scores are used to characterize performance with respect to the localization radius. It can be noted that Figure 4 shows an average result and not all the images will match with its counterpart at a distance of 30-40 meters. It is quite possible to find a match which is as close as 5-10 meters, depending on salient landmarks found in the scene that aid in matching.

## D. Comparative Study

We compare our proposed approach with state-of-the-art place recognition method NetVLAD [1]. We refer to the proposed framework as LoST-X that uses both the LoST descriptor and the keypoint correspondence. The descriptor-only approach is referred to as LoST (that is, without the *keypoint* correspondence and spatial layout verification) in the presented results. All the results on different traverses of both the datasets are generated for front-view vs rear-view matching.

## V. RESULTS

Figures 5 and 6 show the qualitative results using the proposed method. The quantitative results are described in subsequent sections for the proposed pipeline followed by the performance characterization using *single-frame* based matching.

### A. Whole System

Figure 7 and Figure 8 show results for the Oxford and Synthia datasets for sequence-based *multi-frame* matching. It can be observed that the state-of-the-art method does not attain practically usable results even when using sequence-based information, while the proposed system is able to recall a significant fraction of matches without any false positives (e.g. recall at 100% precision).

For the Oxford Robotcar dataset in Figure 7, the proposed system LoST-X consistently shows more than double recall at 100% precision as compared to the other methods. In the most challenging scenario of matching places under simultaneous variation of viewpoint and appearance, the proposed system

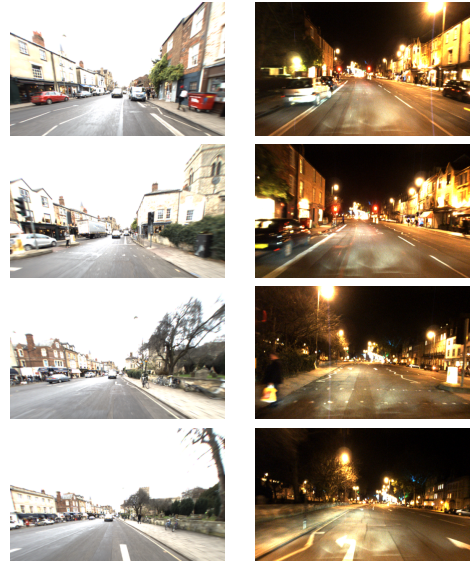

Fig. 5. Matching sequence of images using the proposed system on Oxford Robotcar dataset. The rear-view images are from the Overcast Autumn traverse (left) and front-view images are from the Night Autumn traverse (right).

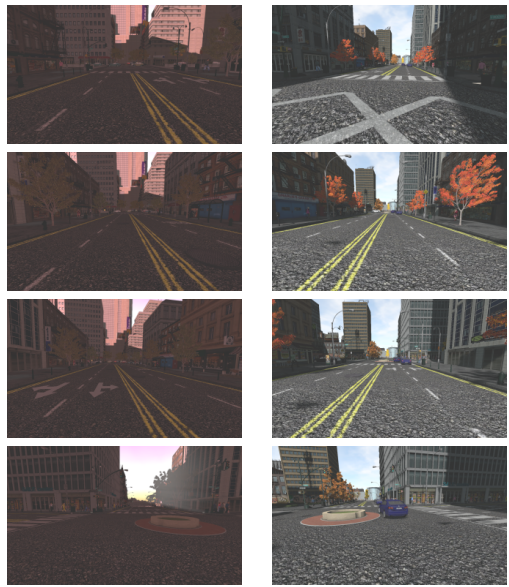

Fig. 6. Matching sequence of images using the proposed system on Synthia Dataset. The rear-view images are from the Dawn traverse (left) and the front-view images are from the Fall traverse (right). The images presented here are brightened by 50% for visualization purposes.

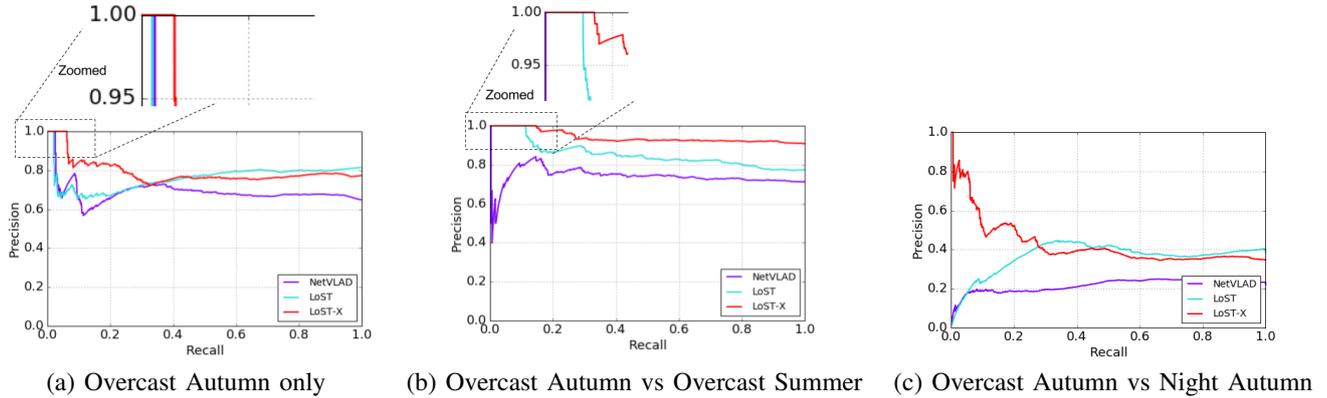

Fig. 7. Oxford Robotcar Dataset: P-R curves for sequence-based *multi-frame* matching of front- and rear-view imagery from different traverses. The proposed method consistently performs better in all the cases exhibiting more than double the recall at 100% precision, especially in the third column with the most challenging scenario of extreme variations in both viewpoint and appearance when other methods never attain 100% precision.

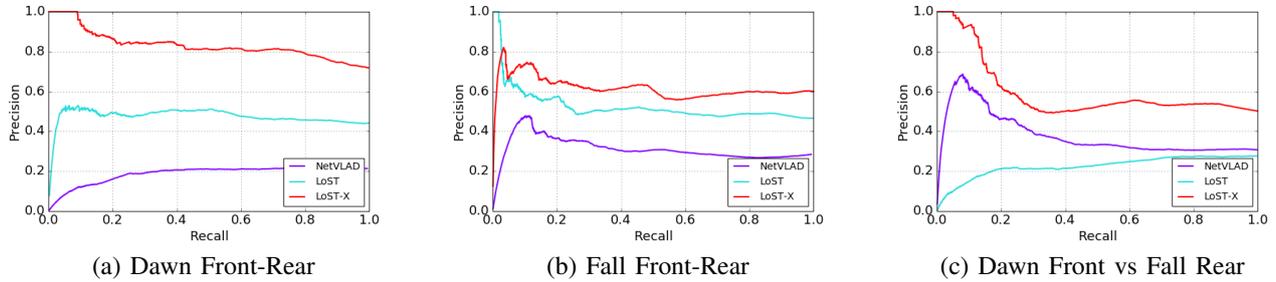

Fig. 8. Synthia Dataset: P-R curves for sequence-based *multi-frame* matching of front- and rear-view imagery from different traverses. The proposed system LoST-X performs significantly better than the baseline method.

continues to perform well, while the baseline never attains 100% precision at any recall level.

Similarly, for the Synthia dataset in Figure 8, LoST-X performs significantly better than the baseline methods, especially in Figure 8 (c) where apart from opposite viewpoints reference and query traverse also exhibit appearance variations due to different times of day and season as also shown in Figure 6. While LoST-X consistently performs the best, both the descriptor-only methods are inconsistent in their relative performance. This is likely attributable to *perceptual aliasing* caused by repeated similar patterns of roads and trees within the traverse and can be avoided by spatial arrangement verification. Furthermore, as shown in Figure 3 (b), there are multiple turns and frequent stops during the traverse. The former is quite challenging because during turns both the front- and rear-view cameras view significant portions of the environment that are not observable by the other camera.

### B. Single-Frame Matching

We established the key results in the previous section using the proposed system and also observed that state-of-the-art method, even with sequence-based information, does not attain practically usable results. Here, we develop further insights and analyze the performance characteristics for *single-frame* based matching.

Figure 9 shows P-R curves and Max-F1 Scores with respect to the localization radius, for comparisons between front- and rear-view images of different traverses of Oxford Dataset. The proposed approach LoST-X consistently performs the best by a significant margin. As compared to the *multi-frame* matching, the performance trends remain similar with respect to the P-R curves. The Max-F1 scores show significantly better performance for lower localization radius and for the extreme scenario of simultaneous viewpoint and appearance variations (Figure 9 (c)) the Max-F1 score is almost double that of the state-of-the-art method at a 30 meter localization radius.

Figure 10 shows performance comparisons for the Synthia Dataset. The Max-F1 scores show that keypoint correspondences significantly improve the performance on top of the descriptor-only matching. The vehicle in this dataset moves through different lanes in a city having similar visual landmarks and some of the intersections are crossed multiple times with varying viewpoint. Therefore spatial arrangement information significantly helps in dealing with perceptual aliasing so caused. The P-R curves for *single-frame* matching have similar trends as compared to *multi-frame* matching, showing that the proposed method (LoST-X) performs consistently better than others.

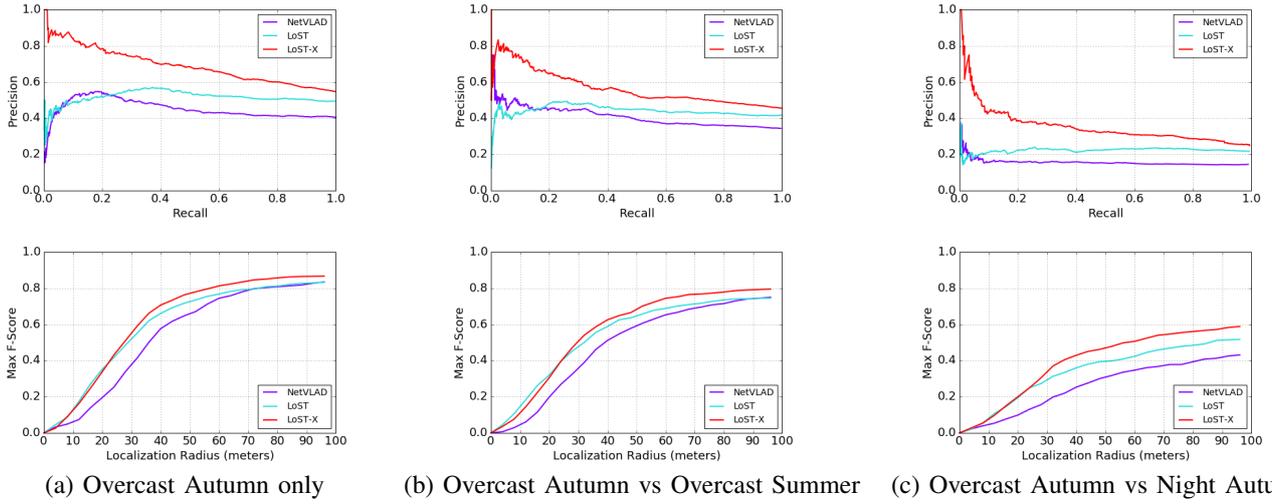

(a) Overcast Autumn only     (b) Overcast Autumn vs Overcast Summer     (c) Overcast Autumn vs Night Autumn

Fig. 9. P-R curves (Top) and Max-F1 Scores (Bottom) for *single-frame* matching between Front- vs Rear-View of different traverses of Oxford Robotcar Dataset. The proposed method consistently performs the best in all cases. In the third column with the most challenging scenario with extreme variations in both viewpoint and appearance, the proposed method has more than double the values of Max-F1 Score for low localization radius.

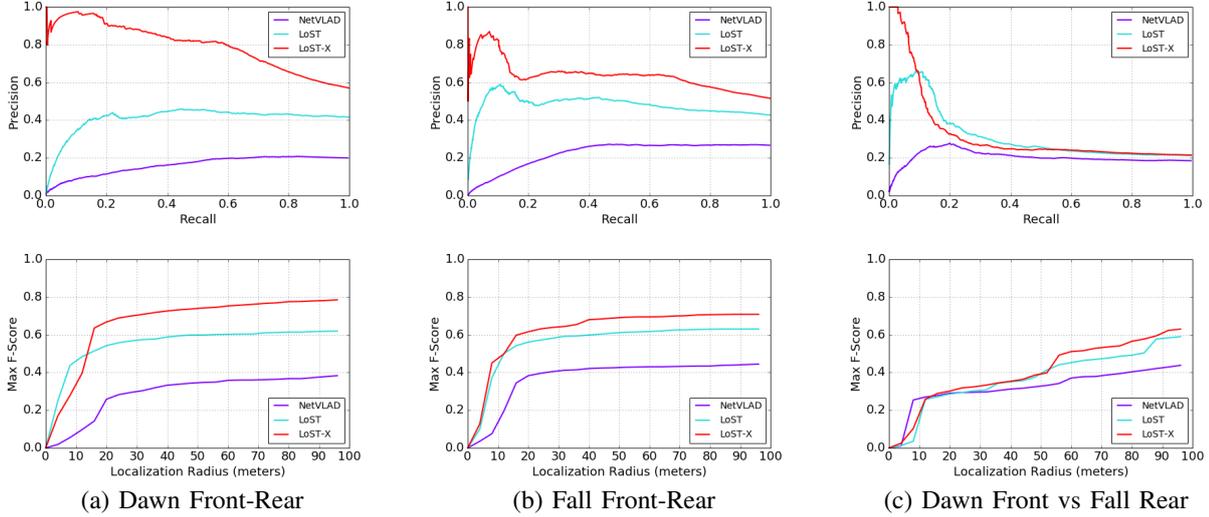

(a) Dawn Front-Rear     (b) Fall Front-Rear     (c) Dawn Front vs Fall Rear

Fig. 10. P-R curves (Top) and Max-F1 Scores (Bottom) for *single-frame* matching between Front- vs Rear-View of different traverses of Synthia Dataset. The spatial consistency check provides a significant gain in performance here by reducing the number of false positives generated due to perceptually similar scenes.

## VI. CONCLUSION

Recognizing a place visually from opposing viewpoints, where only limited parts of the scene are commonly observable from both directions and under extreme appearance change is a very difficult problem. In this research, we presented a place recognition method (LoST-X) that, inspired by the human ability to solve this problem, leverages visual semantic information to effectively describe and match places. We showed through results on two different publicly available datasets that our proposed system can attain a practically usable and significantly higher recall rate at 100% precision than the current state-of-the-art system. We also performed further analysis of performance with respect to *single-frame* based matching and showed that the proposed system consistently performs better than the state-of-the-art in terms of both precision-recall and max-F1 score characteristics. We believe that this is the first time that visual place recognition from opposing viewpoints and with extreme appearance change has been demonstrated without resorting to panoramic sensors or active turn back and look vision techniques.

The use of visual semantics is also appealing in that it may be more readily adaptable to practical applications such as human-robot or human-autonomous vehicle interaction and communication. Critical to this development will be the development of methods that can harness visual semantic information and combine the strengths of both appearance- and geometry-based methods. In this vein, we plan to extend our current work by utilizing semantic blob-level information to match and track individual semantic objects/patches in the environment in order to produce a better scene interpretation. With the availability of sufficient semantically-labeled place recognition data, it may be possible to train an end-to-end system that learns semantic and structural saliency for viewpoint- and appearance-invariant place recognition, by learning to select robust semantic classes with respect to the operating environment and salient *keypoints* for spatial layout consistency. In doing so, we will also gain further insights into the trade offs between a highly engineered, multi-sensory navigation framework such as those prevalent on current autonomous vehicles and primarily vision-driven, more analogous to human navigation approaches.